\documentclass{rescience}

\usepackage{amsfonts}
\usepackage{bm}
\usepackage{siunitx}
\usepackage{booktabs}

\usepackage{xcolor, soul}
\definecolor{mycolor}{RGB}{230,230,255}
\sethlcolor{mycolor}

\usepackage{tabularx}
\usepackage{ragged2e}
\newcolumntype{C}{>{\Centering}X} 
\newcolumntype{L}{>{\RaggedRight}X} 

\usepackage[raggedright]{sidecap} 
\sidecaptionvpos{figure}{c}
\usepackage{tikz}
\usetikzlibrary{arrows}
\usetikzlibrary{shapes}

\begin{document}

\begin{tikzpicture}[remember picture, overlay]
  \node [shape=rectangle, draw=darkred, fill=darkred, yshift=-34mm,
        anchor=north west, minimum width=3.75cm, minimum height=10mm]
        at (current page.north west) {};
  \node [text=white, anchor=center, yshift=-39mm, xshift=1.875cm]
         at (current page.north west) {\small \sffamily RESCIENCE C}; 
\end{tikzpicture}

{\let\newpage\relax\maketitle} \maketitle

\marginnote{
  \footnotesize \sffamily
  \textbf{Edited by}\\
  \ifdefempty{\editorNAME}{\textcolor{darkgray}{(Editor)}}
             {\editorNAME\ifdefempty{\editorORCID}{}{$^{\orcid{\editorORCID}}$}}\\
  ~\\
  \ifdefempty{\reviewerINAME}{}
  {
  \textbf{Reviewed by}\\
  \ifdefempty{\reviewerINAME}{\textcolor{darkgray}{}}
             {\reviewerINAME\ifdefempty{\reviewerIORCID}{}{$^{\orcid{\reviewerIORCID}}$}\\}
  \ifdefempty{\reviewerIINAME}{\textcolor{darkgray}{}}
             {\reviewerIINAME\ifdefempty{\reviewerIIORCID}{}{$^{\orcid{\reviewerIIORCID}}$}\\}
  ~\\
  }
  \textbf{Received}\\
  \ifdefempty{\dateRECEIVED}{---}{\dateRECEIVED}\\
  ~\\
  \textbf{Published}\\
  \ifdefempty{\datePUBLISHED}{---}{\datePUBLISHED}\\
  ~\\
  \textbf{DOI}\\
  \ifdefempty{\articleDOI}{---}{\articleDOI}
}

\newcommand{\container}[1]{\def\@container{#1}}
\begin{container}
  \afterpage {
    \begin{statement}
      \scriptsize \sffamily
      \hrule \vskip .5em
      Copyright © \articleYEAR~\authorsABBRV,
      released under a Creative Commons Attribution 4.0 International license.
      
      Correspondence should be addressed to
      \contactNAME~(\href{mailto:\contactEMAIL}{\contactEMAIL})
  
      The authors have declared that no competing interests exist.

      \ifdefempty{\codeURL}{}
      {Code is available at
      \href{\codeURL}{\detokenize\expandafter{\codeURL}}\ifdefempty{\codeDOI}{.}{ -- DOI \doi{\codeDOI}.}\ifdefempty{\codeSWH}{.}{ -- SWH  \href{https://archive.softwareheritage.org/\codeSWH/}{\detokenize\expandafter{\codeSWH}}.}}

      \ifdefempty{\dataURL}{}
      {Data is available at
      \href{\dataURL}{\detokenize\expandafter{\dataURL}}\ifdefempty{\dataDOI}{.}{ -- DOI \doi{\dataDOI}.}}

      \ifdefempty{\reviewURL}{}
     {Open peer review is available at \href{\reviewURL}{\detokenize\expandafter{\reviewURL}}.}
    \end{statement}
  }
\end{container}



\begin{abstract}
Kalman filters provide a straightforward and interpretable means to estimate hidden or latent variables, and have found numerous applications in control, robotics, signal processing, and machine learning. One such application is neural decoding for neuroprostheses. In 2020, Burkhart et al. thoroughly evaluated their new version of the Kalman filter that leverages Bayes' theorem to improve filter performance for highly non-linear or non-Gaussian observation models. This work provides an open-source Python alternative to the authors' MATLAB algorithm. Specifically, we reproduce their most salient results for neuroscientific contexts and further examine the efficacy of their filter using multiple random seeds and previously unused trials from the authors' dataset. All experiments were performed offline on a single computer.
\end{abstract}

\section{Introduction}

Brain-computer interfaces (BCIs) have long been a subject of science fiction \cite{bci_review}. Detailed communication with a machine through mere thought has become more technologically feasible with time, but still remains infeasible for the general public. However, for certain groups of people, BCIs are a necessary means to circumvent debilitating circumstances. For example, people experiencing quadriplegia or locked-in syndrome have very little means through which to communicate or interact with the outside world, and thus stand to benefit from thought-controlled interfaces through which they can operate robotic limbs or computers \cite{bci_review, braingate}. As another example, people with impaired control or loss of a limb also benefit from robotic prosthetic limbs that can be controlled through thought alone \cite{neuroprosthetics}. In the aforementioned applications of BCIs, one of the key algorithmic challenges is to accurately estimate some relevant aspect of the user's cognition. Specifically, BCIs and neuroprosthetics often seek to decode a quantifiable motor intention signal that can be used to control robots or cursors, such as the velocity of an intended arm, hand, or finger movement \cite{bci_review, braingate}. Such neural decodings are usually made using information from a subset of the user's neurons, made accessible through invasive electrode technologies or through other non-invasive means \cite{progress_in_bci}.
\\
\\
The Kalman filter \cite{kalman} is a common basis upon which practicioners develop BCI decoding methods \cite{pandarinath2018, refit, dlds}. Burkhart et al. (2020) \cite{burkhart_2020} sought to improve Kalman filter performance in neural decoding efforts by developing the Discriminative Kalman Filter (DKF), which leverages Bayes's theorem to facilitate Bayesian filtering in contexts involving highly non-linear or non-Gaussian observations. The authors presented five different experiments verifying the efficacy of the DKF: the first three experiments (4.2--4.4) consisted of intricate toy examples with known observation models, while the final two were BCI-focused experiments with observations consisting of neural recordings. The last two experiments (4.5 and 4.6) are the most salient to BCI applications because the mapping from neural activity to thoughts or intentions is highly nonlinear \cite{brain_dynamics} and usually unknown to practicioners. We chose to solely replicate Experiment 4.5 because it was the most BCI-oriented experiment whose data and code were publicly available, since Experiment 4.6 involved human data.

\section{Kalman Filter Variations}

\tikzset{
    block/.style={
        ellipse,
        draw,
        text centered,
        align=center,
        minimum width=3.5cm,
        minimum height=1.5cm
    },
    connector/.style={
    thick,
    densely dotted,
     -latex
    }
}

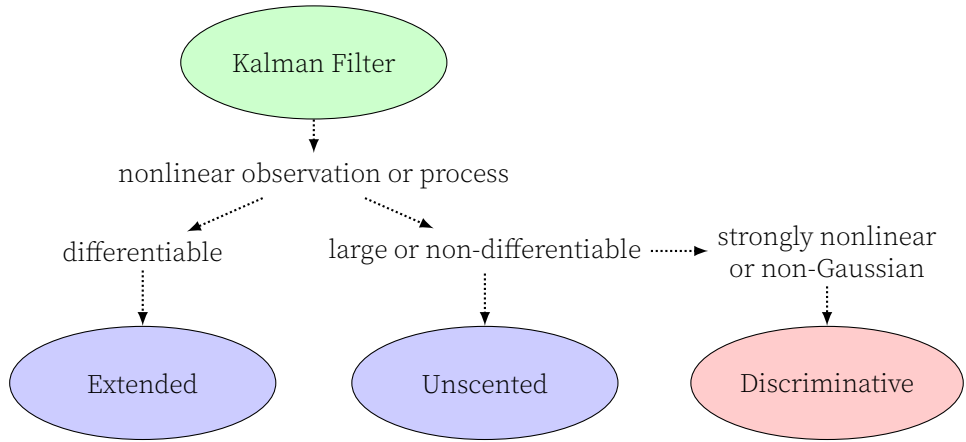
\begin{figure} 
\begin{tikzpicture}

\node [block, fill=green!20] (A) at (2.25,4.25) {Kalman Filter};
\node [block, fill=blue!20] (B) at (0,0) {Extended};
\node [block, fill=blue!20] (C) at (4.5,0) {Unscented};
\node [block, fill=red!20] (D) at (9,0) {Discriminative};

\node [align=center] (txt1) at (2.25, 2.75) {nonlinear observation or process};
\node [align=center] (txt2) at (0, 1.75) {differentiable};
\node [align=center] (txt3) at (4.5, 1.75) {large or non-differentiable};
\node [align=center] (txt4) at (9, 1.75) {strongly nonlinear\\or non-Gaussian};

\draw [connector] (A)->(txt1);
\draw [connector] (txt1)->(txt2);
\draw [connector] (txt1)->(txt3);
\draw [connector] (txt3)->(txt4);

\draw [connector] (txt2)->(B);
\draw [connector] (txt3)->(C);
\draw [connector] (txt4)->(D);
\end{tikzpicture}
\caption{\label{fig:diagram} Summary of how the observation and process models of various Bayesian filtering methods relax those of the Kalman filter.}
\end{figure}

\subsection{Kalman Filter}
Kalman filters \cite{kalman} are a family of algorithms whose purpose is to estimate a set of unobservable latent states $\{\bm{Z}_1,\bm{Z}_2, ..., \bm{Z}_T\}$ given a set of observations $\{\bm{X}_1,\bm{X}_2, ..., \bm{X}_T\}$. Kalman filters operate under the Markov assumption: any observation $\bm{X}_i$ depends only on its corresponding state $\bm{Z}_i$, and any state $\bm{Z}_i$ depends only on the immediately preceding state $\bm{Z}_{i-1}$. Unlike Hidden Markov models \cite{hmm}, which also operate under the Markov assumption, the latent states $\{\bm{Z}_i\}_{i=1}^T$ are not discrete, instead having continuous values. For real-time neural decoding applications, the primary role of Kalman filters is to predict the latest latent state (e.g., a finger velocity) $\bm{Z}_T$ when given the previous latent state $\bm{Z}_{T-1}$ and the latest observation (e.g., neural electrode signals) $\bm{X}_T$. Outside of real-time control, other applications of Kalman filters include smoothing (predicting $\bm{Z}_i$ for any $1 \leq i \leq T$ when given $\{ \bm{X}_i \}_{i=1}^T$) and projection into the future (predicting $\{ \bm{Z}_i \}_{i=T}^\infty$ given $\{ \bm{X}_i \}_{i=1}^T$).
\\
\\
The original Kalman Filter (KF) is a linear, Gaussian, and stationary model, and thus assumes the following:
\begin{enumerate}
    \item Linear Gaussian observations: $p(\bm{X}_i | \bm{Z}_i) \sim \mathcal{N} \left( \bm{HZ}_i, \bm{R} \right)$
    \item Linear Gaussian latent dynamics: $p(\bm{Z}_i | \bm{Z}_{i-1}) \sim \mathcal{N} \left( \bm{AZ}_{i-1}, \bm{G} \right)$\footnote{Some practitioners prefer to write $\bm{A}$ as $\bm{F}$, and $\bm{G}$ as $\bm{Q}$.}
\end{enumerate}

\subsection{Extended and Unscented Kalman Filters}
The KF is the optimal estimator for linear dynamic systems with Gaussian observation and process noise, but neural processing systems are usually highly nonlinear \cite{brain_dynamics}. The simplest modification for handling nonlinear observation models is the Extended Kalman Filter (EKF) \cite{ekf}, which makes the following modifications to the original KF model:
\begin{enumerate}
    \item Nonlinear differentiable Gaussian observations: $p(\bm{Z}_i | \bm{X}_i) \sim \mathcal{N} \left( f^{-1}(\bm{X}_i), \bm{R} \right)$
    \item Nonlinear differentiable Gaussian latent dynamics: $p(\bm{Z}_i | \bm{Z}_{i-1}) \sim \mathcal{N} \left( h(\bm{Z}_{i-1}), \bm{G} \right)$
    \item The observation and process transformations $f^{-1}(\cdot)$ and $h(\cdot)$ cannot be applied directly to the latent state covariance. Instead, their Jacobians evaluated at the current latent value $\bm{Z}_i$ are applied to the latent covariance at timestep $i$.
\end{enumerate}

While the EKF can perform well with sufficient knowledge of the system, it can also perform poorly without such knowledge, or when strong nonlinearities are involved in the system. Another nonlinear Kalman filter algorithm is the Unscented Kalman Filter (UKF) \cite{ukf}. The UKF differs from the EKF by using a deterministic sampling technique known as the unscented transform to pick a minimal set of sample points (sigma points) around the mean. By incorporating sampling techniques, the UKF allows usage of transformations whose Jacobians are difficult or impossible to calculate (i.e., large or non-differentiable functions).


\subsection{Discriminative Kalman Filter}

The Discriminative Kalman Filter (DKF) \cite{burkhart_2016} keeps the following model assumptions from the Kalman Filter:
\begin{enumerate}
    \item Linear Gaussian latent dynamics: $p(\bm{Z}_i | \bm{Z}_{i-1}) \sim \mathcal{N} \left( \bm{AZ}_{i-1}, \bm{G} \right)$ 
    \item The observation model $p( \bm{X}_i | \bm{Z}_i)$ is stationary.
\end{enumerate}


However, unlike the Kalman Filter, the DKF does not assume a linear observation model $p(\bm{X}_i | \bm{Z}_i)$. Instead, the DKF approximates the observation model $p(\bm{X}_i | \bm{Z}_i) \approx p(\bm{Z}_i | \bm{X}_i) / p(\bm{Z}_i)$ via Bayes’ theorem. Such a substitution can prove useful if (a) the observation distribution $p(\bm{X}_i | \bm{Z}_i)$ is strongly nonlinear or non-Gaussian, or (b) the dimensionality of observations $\bm{X}$ is much larger than the dimensionality of latents $\bm{Z}$; in neural signal processing, both are often true. The DKF further models $p(\bm{Z}_i | \bm{X}_i)$ as Gaussian: $p(\bm{Z}_i | \bm{X}_i) \sim \mathcal{N}(f(\bm{X}_i), Q(\bm{X}_i))$, where $f(\cdot)$ and $Q(\cdot)$ are nonlinear functions that map the observation $\bm{X}_i$ to its corresponding elements in the state space $\mathbb{R}^d$ and the covariance space $\mathbb{S}^d$, respectively. Burkhart et al. (2020) \cite{burkhart_2020} formulate that the observation-to-state transformation $f(\cdot)$ and the observation-to-covariance transformation $Q(\cdot)$ are the conditional mean and covariance of $\bm{Z}_i$ given $\bm{X}_i$:
\begin{align}
    f(\bm{X}_i) &= \mathbb{E}(\bm{Z} \, | \, \bm{X} = \bm{X}_i) \label{eq:f_dkf} \\
    Q(\bm{X}_i) &= \mathrm{Var}(\bm{Z} \, | \, \bm{X} = \bm{X}_i) \label{eq:q_dkf}
\end{align}
\\
\\
The DKF also makes the stationarity assumption $p(\bm{Z}_i) \sim \mathcal{N}(\bm{0}, \bm{S})$, where $\bm{S}$ (also written as $\bm{V_Z}$ or $\bm{T}$) is the covariance of $\bm{Z}_i$ when not conditioned on any $\bm{Z}_{i-1}$. Defining the initial latent state estimate $\bm{\mu}_0 = \bm{0}$ and the initial latent covariance estimate $\bm{\Sigma}_0 = \bm{S}$ (the latent covariance $\bm{\Sigma}$ is also written as $\bm{P}$ in traditional filtering literature), each iteration of the DKF algorithm proceeds as follows:
\begin{equation} \label{eq:step1}
    \bm{v}_i = \bm{A \bm{\mu}}_{i-1}
\end{equation}
\begin{equation} \label{eq:step2}
    \bm{M}_i = \bm{A} \bm{\Sigma}_{i-1} \bm{A}^T + \bm{G}
\end{equation}
\begin{equation} \label{eq:step3}
    \bm{\Sigma}_i = \left( \bm{M}_i^{-1} + Q(\bm{X}_i)^{-1} - \bm{S}^{-1} \right)^{-1} 
\end{equation}
\begin{equation} \label{eq:step4}
    \bm{\mu}_i = \bm{\Sigma}_i \left( \bm{M}_i^{-1} \bm{v}_i + Q(\bm{x}_i)^{-1} f(\bm{X}_i) \right)
\end{equation}
\\
\\
In practice, the following additional changes are made to the DKF algorithm, with pseudo-inverses written as $\,^\dagger$:
\begin{enumerate}
    \item $Q(\bm{X}_i)^{-1} - \bm{S}^{-1}$ must be positive definite. If it is not, set $Q(\bm{X}_i)^{-1} = \left( Q(\bm{X}_i)^{-1} + \bm{S}^{-1} \right)^{-1}$.
    \item $ \bm{\Sigma}_i = \left( \bm{M}_i^\dagger + Q(\bm{X}_i)^\dagger - \bm{S}^\dagger \right)^\dagger $
    \item $ \bm{\mu}_i = \bm{\Sigma}_i \left( \bm{M}_i^\dagger \bm{v}_i + Q(\bm{X}_i)^\dagger f(\bm{X}_i) \right)$
\end{enumerate}

Another formulation of the DKF more suitable for real-time applications is the Robust DKF, which makes the assumption that the eigenvalues of $\bm{S}^{-1}$ are so small that the $-\bm{S}^{-1}$ term in Equation \ref{eq:step3} is negligible and can thus be removed. Additionally, the Robust DKF places an improper prior on $\bm{Z}_0$, and modeling starts from $t = 1$. Specifically, $\bm{\mu}_1 = f(\bm{X}_1)$ and $\bm{\Sigma}_1 = Q(\bm{X}_1)$, and iterative calculations start at $t = 2$ instead of $t = 1$.

\section{Methods}

\subsection{Data}

The data that Burkhart et al. (2020) \cite{burkhart_2020} used in their most salient and practical open-source experiment for neuroscientific applications (4.5) came from Flint et al. (2012) \cite{flint_2012}. Specifically, the data is from a 96-channel microelectrode array implanted in the primary motor cortex of one rhesus macaque (Monkey C, center-out from \cite{flint_2012}). The macaque was taught to earn juice rewards by moving a manipulandum in a center-out reaching task. A 128-channel acquisition system recorded the resulting signals, which were sampled at 30 kHz, highpass-filtered at 300 Hz, and then thresholded and sorted into spikes offline. The data is made publicly available by Walker and Kording (2013) \cite{dream} under the Database for Reaching Experiments and Models (DREAM)\footnote{Data is available at \href{https://portal.nersc.gov/project/crcns/download/dream/data_sets/Flint_2012}{https://portal.nersc.gov/project/crcns/download/dream/data\_sets/Flint\_2012}}

\subsection{Preprocessing} \label{preprocessing}

The five \textit{.mat} files from the Flint et al. (2012) manipulandum manipulation trials \cite{flint_2012} are first placed in the same directory as \href{https://github.com/Josuelmet/Discriminative-Kalman-Filter-4.5-Python/blob/main/flint-data-preprocessing/flint_preprocess_data.ipynb}{\textit{flint\_preprocess\_data.ipynb}} for organization and preprocessing. Each file has various recording sessions (henceforth referred to as \textit{trials}), where each session (trial) in a given file was recorded on the same day \cite{flint_2012}. Since the trial data is stored as MATLAB structs, extra attention had to be paid during implementation to ensure that the Python data organization matched the original MATLAB organization method. Each of the five files are processed as follows:

\begin{enumerate}
    \item Isolate all data from the $n$-th trial of the file (some files only have one trial, while other file have up to four trials).
    \item For each trial in the file:
        \subitem a) Stack each timestamp's 3D manipulandum velocity vertically.
        \subitem b) Discretize the activity of each neuron in the trial into bins of $\frac{1}{30 \text{kHz}} \approx$ \SI{33}{\micro\second}.
        \subitem Stack the neurons' spike bins vertically.
    \item Only keep the first two dimensions of the velocities, since the third dimension is not relevant for operation of the manipulandum.
    \item Save the resulting 2D array of manipulandum velocities from the i-th trial as a unique entry in a list of 2D velocity arrays. In equivalent fashion, store the array of spike bins in a list of 2D bin arrays.
    \item Repeat all earlier steps for each trial from each file. There are a total of 12 trials across all files, treating each trial as independent from the others.
\end{enumerate}

Once organized, the lists of trial velocities and spike bins are preprocessed in the same manner as Burkhart et al. (2020) \cite{burkhart_2020}:
\begin{enumerate}
    \item Isolate the velocities and spike bins from the $n$-th trial.
    \item Downsample the spike bin samples from \SI{33}{\micro\second} intervals to \SI{100}{\milli\second} intervals.
    \item Replace the spike bin data with a moving sum (with a window length of 10 entries) of the spike bin data.
    \item Downsample the velocity data samples into intervals of \SI{100}{\milli\second}---specifically, keep entries whose indices are halfway between the indices that were used to downsample the spike bin data earlier.
    \item Replace the spike bin data with the z-scores (using 1 degree of freedom correction, as per MATLAB's zscore function) of its top 10 principal components.
    \item As during organization, store the 2D array of processed spike bins and the 2D array of processed velocities in a list of 2D spike arrays and a list of 2D velocity arrays, respectively.
    \item Repeat all previous steps for each of the 12 trials.
\end{enumerate}

In similar fashion as Burkhart et al. (2020) \cite{burkhart_2020}, data preprocessing yields an array of processed velocities\footnote{\href{https://github.com/Josuelmet/Discriminative-Kalman-Filter-4.5-Python/blob/main/flint-data-preprocessing/procd_velocities.npy}{https://github.com/Josuelmet/Discriminative-Kalman-Filter-4.5-Python/blob/main/flint-data-preprocessing/procd\_velocities.npy}} and an array of processed spike bins\footnote{\href{https://github.com/Josuelmet/Discriminative-Kalman-Filter-4.5-Python/blob/main/flint-data-preprocessing/procd_spikes.npy}{https://github.com/Josuelmet/Discriminative-Kalman-Filter-4.5-Python/blob/main/flint-data-preprocessing/procd\_spikes.npy}}. The resulting latent states (manipulandum velocities) are 2-dimensional, while the observations (z-scored principal component scores of neural activity) are 10-dimensional.

\subsection{Computation}
\href{https://github.com/Josuelmet/Discriminative-Kalman-Filter-4.5-Python/blob/main/Paper_Script_45.ipynb}{\textit{Paper\_Script\_45.ipynb}} reproduces Experiment 4.5 from Burkhart et al. (2020) \cite{burkhart_2020} in Python, based on the authors' original MATLAB implementation. While there are a total of 12 trials, the authors only conducted analysis on the first 6 trials' data. The following analyses and procedures were performed separately on each of the aforementioned trials. See Table \ref{summarized_differences} for an overview of the similarities and differences between our computational implementations and those of Burkhart et al. (2020) \cite{burkhart_2020}.

\begin{table}[ht]
\centering
\begin{tabularx}{\textwidth}{CCC}
\toprule
Algorithm & Burkhart et al. (2020) & Python Reproduction \\
\cmidrule(lr){1-3}
Preprocessing & See Section \ref{preprocessing} & Same, to the best of our ability \\
\midrule
Kalman Filter & Fully deterministic matrix implementation & Same \\
\midrule
Neural Network & One hidden layer of 10 tanh neurons optimizied via Bayesian-regularized Levenberg-Marquardt method & Two hidden layers of 10 tanh neurons optimizied via RMSProp with $l2$ weight penalty and 4,000 epochs \\
\midrule
Nadaraya-Watson &  Kernel bandwidth optimization via the MATLAB fminunc function & Kernel bandwidth optimization via the scipy minimize\_scalar function \\
\midrule
Gaussian Process & Fitting via GPML package with RBF kernel & Fitting via scikitlearn GaussianProcessRegressor with RBF kernel \\
\midrule
Long Short-Term Memory & One 20-dimensional LSTM layer and one fully-connected layer optimized via AdaGrad with dropout & One 20-dimensional LSTM layer and one fully-connected layer optimized via Adam with an $l2$ weight penalty and no dropout \\
\midrule
Extended and Unscented Kalman Filters & Native MATLAB implementation using the earlier one-layer neural network architecture & FilterPy implementation using the earlier two-layer neural network architecture \\
\bottomrule
\end{tabularx}
\caption{The summarized differences between the algorithmic implementations in Burkhart et al. (2020) \cite{burkhart_2020} and this study.}
\label{summarized_differences}
\end{table}

\subsubsection{Training, Validation, and Test Data Split} Before performing any regression, the data from the current trial is isolated and split into training and test data. Recall that each trial represents a recorded session of a macaque center-out reaching task \cite{flint_2012}, and has several thousand \SI{100}{\milli\second} samples (approximately 10 minutes of recorded data) \cite{flint_2012}. The first 5,000 samples are used as training data, while the subsequent 1,000 samples are used as test data. Various methods are used for learning either the observation-to-state transformation $f:\mathbb{R}^{10} \rightarrow \mathbb{R}^2$ or the state-to-observation transformation $f^{-1}:\mathbb{R}^2 \rightarrow \mathbb{R}^{10}$. However, only Nadaraya-Watson (NW) kernel regression \cite{nadaraya1964} is used for learning the transformation $Q:\mathbb{R}^{10} \rightarrow \mathbb{S}^2$ that estimates the conditional covariance of the latent $\bm{Z}_i$ given the observation $\bm{X}_i$. Following Burkhart et al. (2020) \cite{burkhart_2020}, we used 70\% of the training data (3,500 samples) purely to train regression models, while the remaining 30\% (1,500 samples) are used to learn Q(x) using NW regression; thus. The aforementioned 3,5000 samples will be referred to as training data, the next 1,500 samples will be referred to as validation data, and the last 1,000 samples will be referred to as testing data, for clarity. While the indices of the 5,000 (training + validation) and 1,000 (testing) samples are not randomized due to the temporally sensitive nature of BCI decoding, the indices of the 3,500 (training) and the 1,500 (validation) samples are randomly drawn (without replacement) from the 5,000 samples.

\subsubsection{Linear Kalman Filter}
The first regression method is the fundamental baseline upon which to compare all subsequent algorithms: the traditional Kalman filter. It uses all 5,000 training and validation samples as training data, since it does not need a validation set with which to learn a Q(x) covariance function. Aside from estimated latent state means and covariances, the Kalman filter also yields the transition matrix $\bm{A}$, the process noise covariance $\bm{Q}$ (named $\bm{G}$ in the notebook), and the initial estimate covariance $\bm{V_Z}$ (also referred to as $\bm{V}_0$ or $\bm{P}_0$).

\subsubsection{Neural Network Regression}
The next regression method is the Discriminative Kalman Filter (DKF) using neural network (NN) regression. Recall that all DKF methods must learn the observation-to-state transformation $f : \mathbb{R}^{10} \rightarrow \mathbb{R}^2$. Neural network regression estimates $f(\cdot)$ with a feedforward network that learns from the training data (3,500 samples).
\\
\\
Burkhart et al. (2020) \cite{burkhart_2020}  used a neural network with one hidden layer of 10 hyperbolic tangent neurons. However, the authors trained their smaller network using a combination of Levenberg-Marquardt (LM) optimization algorithm and Bayesian regularization (BR) \cite{foresee1997}, which automatically calculate $l2$ weight penalties iteratively and incorporate information from the inverse Hessian of the loss function. However, since resources for second-order optimization are scarce in Python, we used a more traditional neural network architecture---2 hidden layers of 10 hyperbolic tangent neurons each---and optimization method---4,000 epochs using RMSProp \cite{gradient_descent} with a learning rate of $10^{-3}$ and an $l2$ regularization penalty of $10^{-4}$.
\\
\\
After estimating the function $f:\mathbb{R}^{10} \rightarrow \mathbb{R}^2$, the network predicts the latent states (velocities) corresponding to the validation data observations (1,500 samples of processed neural recordings). The optimal bandwidth of the radial basis kernel for covariance estimation is then found by minimizing the leave-one-out mean squared error of Nadaraya-Watson (NW) kernel regression using the validation set and the outer product of the validation residuals ($\bm{Z}_i - f(\bm{X}_i)$ for all $\{\bm{Z}_i, \bm{X}_i\}$ pairs in the validation data). Next, the network predicts the latent states corresponding to the test data observations (1,000 samples), while NW kernel regression predicts the latent state estimate covariances using the test data, validation data, and validation residuals. The final DKF-NN predictions are made by passing the network’s predicted states and covariances, along with the aforementioned Kalman filter parameters $\bm{A}$ (state transition matrix), $\bm{G}$ (the process noise covariance, also written as $\bm{Q}$), and $\bm{V}_Z$ (the stationary covariance of $\bm{Z_i}$ without conditioning on any $\bm{Z}_{i-1}$).

\subsubsection{Nadaraya-Watson Kernel Regression}

Discriminative Kalman filtering via Nadaraya-Watson (NW) kernel regression functions similarly to how NW regression estimated the state covariance function $Q:\mathbb{R}^{10} \rightarrow \mathbb{S}^2$ in the case of neural network regression. First, the bandwidth of the radial basis kernel is optimized to minimize the leave-one out mean squared error in the estimated function $f:\mathbb{R}^{10} \rightarrow \mathbb{R}^2$ using the (3,500) training samples. Next, $f(\cdot)$ predicts the latent states (velocities) of the (1,5000) validation samples. $Q(\cdot)$ is then estimated from the resulting validation residuals in the same fashion as in neural network regression. For the (1,000) test sample observations, $f(\cdot)$ and $Q(\cdot)$ then predict the latent states and state covariances, which are processed into the final DKF-NW estimates in the same manner as in the neural network regression case.

\subsubsection{Gaussian Process Regression}
The final filtering method dependent on the Discriminative Kalman Filter involves estimating $f:\mathbb{R}^{10} \rightarrow \mathbb{R}^2$ via Gaussian process (GP) regression \cite{gpml}. As with neural network and Nadaraya-Watson (NW) regression, the residuals of the estimated function $f(\cdot)$ are calculated on the validation data and then used to estimate $Q(\cdot)$. Afterwards, $f(\cdot)$ and $Q(\cdot)$ predict the latent states and covariances of the test data. The predictions undergo the DKF algorithm to produce a final DKF-GP estimate in the same manner as earlier DKF regressions.
\\
\\
The authors utilized the GPML MATLAB package to train their Gaussian process models \cite{gpml, rasmussen2010}. However, due to the lack of modern Python ports of GPML, we instead used the GaussianProcessRegressor (GPR) class from scikit-learn\footnote{\href{https://scikit-learn.org/stable/modules/generated/sklearn.gaussian_process.GaussianProcessRegressor.html}{sklearn.gaussian\_process.GaussianProcessRegressor}}. While the GPR class and GPML package both use algorithms from the same work \cite{gpml}, GPR regression is less customizable and flexible than GPML regression, resulting in worse accuracy (but improved runtime) compared to Burkhart et al. (2020) \cite{burkhart_2020}, despite our usage of the same kernel type (radial basis function / squared exponential). Unlike the authors, who use two separate $\mathbb{R}^{10} \rightarrow \mathbb{R}$ Gaussian process regressions to compose $f:\mathbb{R}^{10} \rightarrow \mathbb{R}^2$, we use a single $\mathbb{R}^{10} \rightarrow \mathbb{R}^2$ Gaussian process regression, since we did not observe any improvements in performance from using the former method (likely due in part to our usage of an isotropic kernel). We also did not find any improvements from using an anisotropic kernel.

\subsubsection{Long Short-Term Memory Regression}
Unlike all other regression methods, Long Short-Term Memory (LSTM) regression \cite{lstm} does not use any explicit Kalman filtering framework and does not learn $f(\cdot)$ in the same manner as described in Equation \ref{eq:f_dkf}. Instead, we and Burkhart et al. (2020) \cite{burkhart_2020} used an LSTM recurrent network that learns $f(\cdot)$ conditioned on $\bm{X}_i$ and its previous two observations $\bm{X}_{i-1}$ and $\bm{X}_{i-2}$.
\\
\\
We were successfully able to replicate the authors' LSTM network, since they also wrote theirs in Python. However, because we used modern PyTorch while they used the older TensorFlow V1 framework, we still had to translate their architectural methods to our newer framework. In accordance with the authors' work, our model consisted of one LSTM layer of hidden dimensionality 20 that recurrently processes 3 observations from $\mathbb{R}^{10}$ before its hidden state undergoes a linear projection onto $\mathbb{R}^{20}$. Unlike the authors, we used Adam optimization (with a learning rate of $10^{-3}$ and an $l2$ weight penalty of $10^{-4}$) and no dropout, since the LSTM performed better with those adjustments made. Note that the LSTM had significantly fewer parameters than data points, which is likely why dropout did not improve performance.
\\
\\
Unlike the DKF, EKF, and UKF methods, LSTM regression partitions the training and validation differently. Recall that there are 5,000 observations in the training and validation data. The LSTM training and validation data are not drawn randomly; i.e., the first 3,500 observations are for training, while the last 1,500 are for validation.
\\
\\
As in all previous methods, $Q:\mathbb{R}^{10} \rightarrow \mathbb{S}^2$ is learned via Nadaraya-Watson kernel regression. Interestingly, Burkhart et al. (2020) did not apply the DKF to process LSTM-estimated states. Upon further investigation, we found that their choice made empirical sense, since DKF processing worsened LSTM performance, as can be seen in  Tables \ref{tab:nrmse-all-trials} and \ref{tab:maae-all-trials}. In order to further investigate the interactions between sequence models and DKF methods, we also tried using a Transformer \cite{attention} model to estimate $f(\bm{X}_i, \bm{X}_{i-1}, \bm{X}_{i-2})$. However, given our small input timestep size of 3, we found that the LSTM architecture had superior performance in both the normalized root mean square error (nRMSE) and mean absolute angle error (MAAE) metrics with which we valuate filtering methods, required at least 10x fewer parameters, and was easier to train.

\subsubsection{Extended and Unscented Kalman Filters}

Burkhart et al. (2020) \cite{burkhart_2020} used the existing native MATLAB implementations of the Extended and Unscented Kalman filters, while we used the FilterPy library \cite{filterpy}.
Usage of the Extended Kalman Filter (EKF) \cite{ekf} and Unscented Kalman Filter (EKF) \cite{ukf} begins with learning the state-to-observation function $f^{-1}:\mathbb{R}^2 \rightarrow \mathbb{R}^{10}$, unlike the previous DKF methods. To learn $f^{-1}(\cdot)$, we used a neural network with the same architecture (albeit with flipped input and output dimensionalities) as that of the DKF-NN method (2 hidden layers of 10 hyperbolic tangent neurons), since the authors' original $f^{-1}(\cdot)$ network faced the same issues of reproduction in Python as their $f(\cdot)$ network. We also used the same optimization method, hyperparameters, and training data as in the DKF-NN method (4000 iterations, $10^{-3}$ learning rate, and a $10^{-4}$ $l2$ weight penalty). For both the EKF and the UKF, the observation noise $\bm{R}$ is estimated as the covariance of the residuals evaluated over the validation data (i.e., the covariance of $\bm{X}_i - f(\bm{Z}_i)$ for all $\{\bm{X}_i, \bm{Z}_i\}$ pairs in the validation data).
\\
\\
The EKF uses the learned $f^{-1}(\cdot)$ as its observation function. The Jacobian of the function is available through PyTorch, and is supplied to the EKF algorithm. Along with the aforementioned information from $f^{-1}(\cdot)$, the EKF uses the residual-estimated $\bm{R}$ and the Kalman filter parameters $\bm{A}$ (the state transition matrix), $\bm{G}$ (the process noise, also written as $\bm{Q}$), and an initial covariance estimate $\bm{P}_0 = \bm{V}_Z$ to iteratively calculate predictions over the test data.
\\
\\
The UKF uses the same observation function $f^{-1}(\cdot)$ as the EKF. While the UKF can operate with explicitly nonlinear state transitions $F:\mathbb{R}^2 \rightarrow \mathbb{R}^2$ (unlike the EKF), the state transition function used here (in accordance with the authors' work) is set as multiplication by the Kalman state transition matrix: $F(\bm{Z_i}) = \bm{AZ}_i$. The UKF uses the same Kalman parameters as the EKF ($\bm{G}$, $\bm{R}$, and $\bm{P}_0 = \bm{V}_Z$) and generates predictions over the test data in the same fashion as the EKF. One key difference between the UKF algorithm here and the UKF algorithm used by Burkhart et al. (2020) \cite{burkhart_2020} is that MATLAB differs from FilterPy in how the number of sample (sigma) points are calculated.








\section{Results}

\renewcommand{\bf}{\textbf}

\begin{table}[ht]
\centering \scriptsize
\begin{tabular}[t]{lccccccc}
\toprule
         & Trial 1 & Trial 2 & Trial 3 & Trial 4 & Trial 5 & Trial 6 & Average \\
\cmidrule(lr){2-8} 
Kalman   & 0.765 | \bf{0.765} & 0.945 | \bf{0.942} & 0.788 | \bf{0.788} & 0.792 | \bf{0.793} & 0.779 | \bf{0.780} & 0.761 | \bf{0.765} & 0.805 | \bf{0.805} \\
\midrule
DKF-NW   & --19\% | \bf{--21\%} & \hl{--18\%} | \bf{--18\%} & \enspace --9\% | \bf{--17\%} & \hl{--21\%} | \bf{--23\%} & --19\% | \bf{--20\%} & --20\% | \bf{--23\%} & \hl{--18\%} | \bf{--20\%} \\
DKF-GP   & \enspace --7\% | \bf{--21\%} & --11\% | \bf{--19\%} & \enspace --8\% | \bf{--15\%} & \enspace --9\% | \bf{--20\%} & --12\% | \bf{--18\%} & \enspace --9\% | \bf{--20\%} & \enspace --9\% | \bf{--19\%} \\
DKF-NN   & \hl{--23\%} | \bf{--19\%} & \enspace \enspace 0\% | \bf{--15\%} & \enspace --7\% | \bf{--13\%} & --15\% | \bf{--13\%} & --19\% | \bf{--13\%} & --22\% | \bf{--17\%} & --14\% | \bf{--15\%} \\
LSTM     & --22\% | \bf{--15\%} & \enspace --1\% | \bf{--19\%} & \hl{--21\%} | \bf{--16\%} & --19\% | \bf{--13\%} & \hl{--21\%} | \bf{--16\%} & \hl{--25\%} | \bf{--11\%} & \hl{--18\%} | \bf{--15\%} \\
EKF      & --1\% | \bf{2\%} \,\, & \enspace 7\% | \bf{24\%} & \enspace 8\% | \bf{12\%} & 18\% | \bf{18\%} & 11\% | \bf{12\%} & 6\% | \bf{3\%} & \enspace 8\% | \bf{12\%} \\
UKF      & 1\% | \bf{2\%} & \enspace 1\% | \bf{31\%} & \enspace 6\% | \bf{18\%} & 11\% | \bf{18\%} & \enspace 5\% | \bf{15\%} & 3\% | \bf{6\%} & \enspace 4\% | \bf{15\%} \\
\bottomrule
\end{tabular}
\caption{Normalized RMSE (nRMSE) between the predicted and true test velocities using one random seed, as in Table 1 from Burkhart et al. (2020) \cite{burkhart_2020}. Bolded values indicate the authors' published results, while highlighted values denote our best result for each trial. Note that predicting identically zero would yield a nRMSE of 1.}
\label{tab:nrmse}
\end{table}

\begin{table}[ht]
\centering \scriptsize
\begin{tabular}[t]{lccccccc}
\toprule
         & Trial 1 & Trial 2 & Trial 3 & Trial 4 & Trial 5 & Trial 6 & Average \\
\cmidrule(lr){2-8}
Kalman   & 0.884 | \bf{0.889} & 0.957 | \bf{0.955} & 1.026 | \bf{1.025} & 0.930 | \bf{0.933} & 0.966 | \bf{0.964} & 0.926 | \bf{0.926} & 0.948 | \bf{0.949} \\
\midrule
DKF-NW   & \hl{--14\%} | \bf{--15\%} & \enspace \hl{0\%} | \bf{--1\%} & \hl{--21\%} | \bf{--20\%} & \hl{--15\%} | \bf{--17\%} & \hl{--25\%} | \bf{--25\%} & \hl{--29\%} | \bf{--28\%} & \hl{--17\%} | \bf{--18\%} \\
DKF-GP   & \enspace --5\% | \bf{--11\%} & 10\% | \bf{7\%} \,\, & --19\% | \bf{--22\%} & \enspace --7\% | \bf{--16\%} & --17\% | \bf{--24\%} & --21\% | \bf{--25\%} & --10\% | \bf{--15\%} \\
DKF-NN   & --9\% | \bf{--7\%} & 11\% | \bf{--2\%} & --15\% | \bf{--17\%} & --11\% | \bf{--16\%} & --19\% | \bf{--21\%} & --21\% | \bf{--23\%} & --11\% | \bf{--14\%} \\
LSTM     & --6\% | \bf{--2\%} & 13\% | \bf{--2\%} & --18\% | \bf{--12\%} & --12\% | \bf{--6\%} \, & --20\% | \bf{--10\%} & --20\% | \bf{--8\%} \,\, & --11\% | \bf{--7\%} \,\, \\
EKF      & 1\% | \bf{4\%} & 12\% | \bf{3\%} \, & \enspace 2\% | \bf{--2\%} & \enspace 5\% | \bf{--4\%} & --8\% | \bf{--8\%} & --1\% | \bf{--7\%} & \enspace 2\% | \bf{--2\%} \\
UKF      & --1\% | \bf{0\%} \,\, & 9\% | \bf{3\%} & \enspace 1\% | \bf{--3\%} & \enspace 2\% | \bf{--3\%} & --10\% | \bf{--8\%} \,\, & --6\% | \bf{--6\%} & --1\% | \bf{--3\%} \\
\bottomrule
\end{tabular}
\caption{Mean Absolute Angle Error (MAAE, in radians) of the predicted and true test velocities using one random seed, as in Table 2 from Burkhart et al. (2020) \cite{burkhart_2020}. Bolded values indicate the authors' published results, while highlighted values denote our best result for each trial. Note that chance prediction would yield a MAAE of $\pi/2 \approx 1.57$ radians. The authors argued that MAAE may be a more salient metric for neuroprosthetic applications.}
\label{tab:maae}
\end{table}

After reproducing the regression methods implemented by Burkhart et al. (2020) \cite{burkhart_2020}, we evaluated their performance on the held-out 1,000 samples of test data of the first six of twelve trials, in the same manner as the original authors and with the same random seed. Specifically, we calculated the normalized root-mean square error (nRMSE) and mean absolute angle error (MAAE) between the predicted and ground truth test data velocities, as shown in Tables \ref{tab:nrmse} and \ref{tab:maae}, respectively. Bolded values indicate the authors' original results. 
While we were unable to replicate most of the exact numerical results from Burkhart et al. (2020) \cite{burkhart_2020}, we faithfully reproduced the trends in performance from the authors' work. Notable exceptions include worsened DKF-GP performance (due to the difficulty of translating GPML computations to Python) and heightened LSTM performance (due to added hyperparameter tuning). Although the DKF-NW and LSTM regression methods had the same average performance over the six trials, our results agreed with those of Burkhart et al. (2020) \cite{burkhart_2020} in that the DKF-NW triumphed as the best regression method due to its superior performance with respect to the MAAE metric.

\begin{table}[ht]
\centering \scriptsize
\begin{tabular}[t]{lrrrrrrrrr}
\toprule
          & Trial 1 & Trial 2 & Trial 3 & Trial 4 & Trial 5 & Trial 6 & Trial 9 & Trial 10 & Average \\
\cmidrule(lr){2-10}
K   & 0.76 & 0.94 & 0.79 & 0.79 & 0.78 & 0.76 & 1.01 & 0.92 & 0.84 \\
\midrule
NW        & -15\% & \hl{-18\%} & -18\% & \hl{-24\%} & -20\% & -21\% & -5\% & 3\% & -15\% \\
DKF-NW   & -18\% & \hl{-18\%} & -12\% & -19\% & -18\% & -20\% & -5\% & 2\% & -14\% \\
\midrule
GP        & -3\% & -9\% & -7\% & -8\% & -10\% & -6\% & -7\% & -14\% & -8\% \\
DKF-GP   & -6\% & -11\% & -8\% & -8\% & -12\% & -8\% & \hl{-8\%} & \hl{-16\%} & -10\% \\
\midrule
NN        & -19\% & -17\% & -21\% & -23\% & -20\% & -23\% & -5\% & -7\% & \hl{-17\%} \\
DKF-NN   & -20\% & -11\% & -12\% & -16\% & -17\% & -20\% & -4\% & -9\% & -14\% \\
\midrule
LSTM      & \hl{-22\%} & 12\% & \hl{-23\%} & -21\% & \hl{-23\%} & \hl{-25\%} & -7\% & -12\% & -15\% \\
DKF-LSTM & -15\% & 84\% & -8\% & -7\% & -15\% & -14\% & -4\% & -12\% & 1\% \\
\midrule
EKF       & 2\% & 10\% & 11\% & 18\% & 14\% & 8\% & 5\% & 12\% & 10\% \\
UKF       & -1\% & -1\% & 6\% & 15\% & 10\% & 4\% & -5\% & 2\% & 4\% \\
\bottomrule
\end{tabular}
\caption{Average Normalized RMSE (nRMSE) between the predicted and true test velocities, with and without DKF filtering, evaluated over ten different random seeds. The best results from each trial are highlighted. Trials 7, 8, 11, and 12 did not have the requisite number of samples and were thus not included.}
\label{tab:nrmse-all-trials}
\end{table}

\begin{table}[ht]
\centering \scriptsize
\begin{tabular}[t]{lrrrrrrrrr}
\toprule
          & Trial 1 & Trial 2 & Trial 3 & Trial 4 & Trial 5 & Trial 6 & Trial 9 & Trial 10 & Average \\
\cmidrule(lr){2-10}
K & 0.88 & 0.96 & 1.03 & 0.93 & 0.97 & 0.93 & 0.97 & 1.03 & 0.96 \\
\midrule
NW       & -8\% & 1\% & \hl{-21\%} & -15\% & -21\% & -25\% & -1\% & -6\% & -12\% \\
DKF-NW   & \hl{-14\%} & \hl{-1\%} & \hl{-21\%} & \hl{-16\%} & \hl{-25\%} & \hl{-28\%} & \hl{-5\%} & -5\% & \hl{-14\%} \\
\midrule
GP       & 2\% & 12\% & -18\% & -3\% & -13\% & -16\% & 6\% & -4\% & -4\% \\
DKF-GP   & -5\% & 11\% & -19\% & -7\% & -19\% & -20\% & 0\% & \hl{-8\%} & -8\% \\
\midrule
NN       & 1\% & 3\% & -17\% & -7\% & -15\% & -15\% & 4\% & 3\% & -5\% \\
DKF-NN   & -6\% & 2\% & -17\% & -10\% & -19\% & -20\% & -3\% & 1\% & -9\% \\
\midrule
LSTM     & -6\% & 15\% & -20\% & -13\% & -20\% & -20\% & -3\% & -5\% & -9\% \\
DKF-LSTM & -6\% & 24\% & -18\% & -12\% & -20\% & -20\% & -2\% & -5\% & -7\% \\
\midrule
EKF      & 0\% & 8\% & 1\% & 1\% & -4\% & -2\% & 3\% & 19\% & 3\% \\
UKF      & -4\% & 6\% & 0\% & 0\% & -8\% & -7\% & -1\% & 13\% & 0\% \\
\bottomrule
\end{tabular}
\caption{Average Mean Absolute Angle Error (MAAE) between the predicted and true test velocities, with and without DKF filtering, evaluated over ten different random seeds, as in Table \ref{tab:nrmse-all-trials}. The best results from each trial are highlighted.}
\label{tab:maae-all-trials}
\end{table}

To bring further insight into the performance of the Discriminative Kalman filter in neuroscientific contexts, we evaluated the aforementioned regression methods with and without DKF filtering applied. Specifically, we measured their average performance (see Tables \ref{tab:nrmse-all-trials} and \ref{tab:maae-all-trials}) on all eight of the twelve trials that had the requisite number of samples (at least 6,000) using ten different random seeds. To our surprise, we found that unfiltered neural network regression best minimized nRMSE, but that the DKF-NW had the lowest MAAE. Interestingly, DKF application for neural networks increases nRMSE while decreasing MAAE.

\section{Conclusion}

This partial replication study confirms the most salient results from Burkhart et al. (2020) \cite{burkhart_2020} concerning the performance of their Discriminative Kalman filter (DKF) on publicly available neuroprosthetic data. We not only successfully affirmed that the DKF-NW had the best overall performance, but we also conducted further tests that prove the efficacy of the DKF while showing the differences that may occur between different metrics for evaluating filter performance.
\\
\\
While DKF methods improved over the Kalman baselines, the closeness in performance of Kalman and DKF methods to the trivial baseline of predicting identically zero, especially when using the nRMSE metric, indicates that future work is needed to improve such filters. For example, future endeavors could include Discriminative Kalman Filter incorporation into more modern approaches in neural latent state estimation, such as sequential/dynamical autoencoders, modern state-space models, Kalman networks, or transformers \cite{pandarinath2018, dynamical_vaes, dyer_2022}. While we did not find our transformer architecture outperformed LSTMs, it is certainly possible that such an architecture could surpass LSTM performance, especially in trials of much longer durations with longer-range dependencies. Additionally, the effects of DKF application on neural network or LSTM regression are not entirely clear, and could stand to be elucidated in future works.

\section{Acknowledgements}

We give thanks to Flint et al. (2012) \cite{flint_2012} and Walker and Kording (2013) \cite{dream} for collecting and hosting the primate reach data, Burkhart et al. (2020) \cite{burkhart_2020} for their mathematical insights into Bayesian filtering, and Caleb Kemere and Richard G. Baraniuk for discussion of this endeavor.

\hypersetup{linkcolor=black,urlcolor=darkgray}
\renewcommand\emph[1]{{\bfseries #1}}
\setlength\bibitemsep{0pt}
\printbibliography

\end{document}